\documentclass[11pt]{article}
\usepackage{acl2012-onecolumn}
\usepackage{times}
\usepackage{url}
\usepackage{latexsym}
\usepackage{amssymb}
\usepackage{amsthm}
\usepackage{color}

\usepackage{amsmath}
\usepackage{verbatim}
\usepackage{tikz-qtree}
\usepackage{tikz}
\usetikzlibrary{shapes.geometric}
\usetikzlibrary{shapes.arrows}
\usetikzlibrary{patterns}
\usetikzlibrary{chains}
\usetikzlibrary{calc}

\newcommand{\ignore}[1]{}

\newsavebox{\one}
\newsavebox{\two}
\newsavebox{\three}
\newsavebox{\four}
\newsavebox{\five}

\title{Notes on Noise Contrastive Estimation and Negative Sampling}

\author{Chris Dyer \\
  School of Computer Science \\
  Carnegie Mellon University \\
  5000 Forbes Ave., Pittsburgh, PA, 15213 \\
  {\tt cdyer@cs.cmu.edu} \\}
 
\date{}

\begin{document}
\maketitle
\begin{abstract}
Estimating the parameters of probabilistic models of language such as maxent models and probabilistic neural models is computationally difficult since it involves evaluating partition functions by summing over an entire vocabulary, which may be millions of word types in size. Two closely related strategies---\textbf{noise contrastive estimation} \cite{mnih:2012,mnih:2013,vaswani:2013} and \textbf{negative sampling} \cite{mikolov:2012,goldberg:2014}---have emerged as popular solutions to this computational problem, but some confusion remains as to which is more appropriate and when. This document explicates their relationships to each other and to other estimation techniques. The analysis shows that, although they are superficially similar, NCE is a general parameter estimation technique that is asymptotically unbiased, while negative sampling is best understood as a family of binary classification models that are useful for learning word representations but not as a general-purpose estimator.
\end{abstract}

\section{Introduction}
Let us assume the following model of language which predicts a word $w$ in a vocabulary $V$ based on some given context $c$:\footnote{By \emph{language model} I mean a model that generates one word at a time, conditional on any other ambient context such as previously generated or surrounding words, a topic label, text in another language, etc. Excluded are so-called ``whole-sentence'' or ``globally normalized'' language models. While these can also, in principle, be learned using the techniques described in these notes, this exposition focuses on models that predict a single word at a time.}
\begin{align}
p_{\theta}(w \mid c) &=\frac{u_\theta(w,c)}{\sum_{w' \in V} u_\theta(w',c)} = \frac{u_\theta(w,c)}{Z_\theta(c)}, \label{eq:lm}
\end{align}
where $u_{\theta}(w,c) = \exp s_{\theta}(w,c)$ assigns a score to a word in context,  $Z(c)$ is the partition function that normalizes this into a probability distribution, and $s_{\theta}(w,c)$ is differentiable with respect to $\theta$. The standard learning procedure is to maximize the likelihood of a sample of training data. Unfortunately, computing this probability (and its derivatives) is expensive since this requires summing over all words in $V$, which is generally very large.

What can be done? Since the derivatives of the log likelihood include terms that are the expected values of the parameters under the model distribution, the classic strategy has been to use \textbf{importance sampling} and related Monte Carlo techniques to approximate these expectations \cite{bengio:2003}. Noise contrastive estimation and negative sampling represent an evolution of these techniques. These work by transforming the computationally expensive learning problem into a binary classification \emph{proxy problem} that uses the same parameters but requires statistics that are easier to compute.

\subsection{Empirical distributions, Noise distributions, and Model distributions}
I will refer to $\tilde{p}(w \mid c)$ and $\tilde{p}(c)$ as empirical distributions.  Our task is to find the parameters $\theta$ of a model $p_{\theta}(w \mid c)$ that approximates the empirical distribution as closely as possible, in terms of minimal cross-entropy. To avoid costly summations, a ``noise'' distribution, $q(w)$, is used. In practice $q$ is a uniform, empirical unigram, or ``flattened'' empirical unigram distribution (by exponentiating each probability by $0 < \alpha < 1$ and renormalizing).

\section{Noise contrastive estimation (NCE)}

NCE reduces the language model estimation problem to the problem of estimating the parameters of a probabilistic binary classifier that uses the same parameters to distinguish samples from the empirical distribution from samples generated by the noise distribution \cite{gutmann:2010}. The two-class training data is generated as follows: sample a $c$ from $\tilde{p}(c)$, then sample one ``true'' sample from $\tilde{p}(w \mid c)$, with auxiliary label $D=1$ indicating the datapoint is drawn from the true distribution, and $k$ ``noise'' samples from $q$, with auxiliary label $D=0$ indicating these data points are noise. Thus, given $c$, the joint probability of $(d,w)$ in the two-class data has the form of the mixture of two distributions:
\begin{align*}
p(d, w \mid c) &= \begin{cases} \frac{k}{1+k} \times q(w) & \textrm{if }d=0 \\
\frac{1}{1+k} \times \tilde{p}(w \mid c) & \textrm{if }d=1
\end{cases}.
\end{align*}
Using the definition of conditional probability, this can be turned into a conditional probability of $d$ having observed $w$ and $c$:
\begin{align*}
p(D = 0 \mid c,w) &=\frac{\frac{k}{1+k} \times q(w)}{\frac{1}{1+k} \times  \tilde{p}(w \mid c)  + \frac{k}{1+k} \times q(w)} \\
&= \frac{k \times q(w)}{ \tilde{p}(w \mid c) + k \times q(w)}\\
p(D = 1 \mid c,w) &= \frac{ \tilde{p}(w \mid c) }{ \tilde{p}(w \mid c)  + k \times q(w)}.
\end{align*}
Note that these probabilities are written in terms of the empirical distribution.

NCE replaces the empirical distribution $\tilde{p}(w \mid c)$ with the model distribution $p_{\theta}(w \mid c)$, and $\theta$ is chosen to maximize the conditional likelihood of the ``proxy corpus'' created as described above. But, thus far, we have not solved any computational problem yet: $p_{\theta}(w \mid c)$ still requires evaluating the partition function---all we have done is transform the objective by adding some noise. To avoid the expense of evaluating the partition function, NCE makes two further assumptions. First, it proposes partition function value $Z(c)$ be estimated as parameter $z_c$ (thus, for every empirical $c$, classic NCE introduces one parameter). Second, for neural networks with lots of parameters, it turns out that fixing $z_c=1$ for all $c$ is effective \cite{mnih:2012}. This latter assumption both reduces the number of parameters and encourages the model to have ``self-normalized'' outputs (i.e., $Z(c) \approx 1$). Making these assumptions, we can now write the conditional likelihood of being a noise sample or true distribution sample in terms of $\theta$ as
\begin{align*}
p(D = 0 \mid c,w) &=\frac{k \times q(w)}{u_\theta(w,c) + k \times q(w)} \\
p(D = 1 \mid c,w) &=\frac{ u_{\theta}(w,c) }{u_\theta(w,c) + k \times q(w)}.
\end{align*}
We now have a binary classification problem with parameters $\theta$ that can be trained to maximize conditional log-likelihood of $\mathcal{D}$, with $k$ negative samples chosen:
\begin{align*}
\mathcal{L}_{\textsc{nce}_k} &= \sum_{(w,c) \in \mathcal{D}} \left( \log p(D=1 \mid c,w) + k \mathbb{E}_{\overline{w} \sim q} \log p(D = 0 \mid c, \overline{w}) \right).
\end{align*}
Unfortunately, the expectation of the second term in this summation is still a difficult summation---it is $k$ times the expected log probability (according to the current model) of producing a negative label under the noise distribution over all words in $V$ in a context $c$. We still have a loop over the entire vocabulary. The final step is therefore to replace this expectation with its Monte Carlo approximation:
\begin{align*}
\mathcal{L}^{\textsc{mc}}_{\textsc{nce}_k} &= \sum_{(w,c) \in \mathcal{D}} \left( \log p(D=1 \mid c,w) +  k \times \sum^k_{i=1,\overline{w} \sim q} \frac{1}{k} \times \log p(D = 0 \mid c, \overline{w}) \right) \\
 &= \sum_{(w,c) \in \mathcal{D}} \left( \log p(D=1 \mid c,w) +   \sum^k_{i=1,\overline{w} \sim q}  \log p(D = 0 \mid c, \overline{w}) \right).
\end{align*}

\subsection{Asymptotic analysis}
Although the objective $\mathcal{L}_{\textsc{nce}_k}$ is intractable, its derivative sheds light on why NCE works. This quantity may be written as
\begin{align*}
\frac{\partial}{\partial \theta} \mathcal{L}_{\textsc{nce}_k} =  \sum_{(w',c) \in \mathcal{D}}  \left( \sum_{w \in V}\frac{k \times q(w)}{u_{\theta}(w \mid c) + k \times q(w)} \times \left(\tilde{p}(w \mid c) - u_{\theta}(w \mid c) \right) \frac{\partial}{\partial \theta} \log u_{\theta}(w \mid c) \right).
\end{align*}
It is easy to see that in the limiting case as $k \rightarrow \infty$, this derivative tends to the gradient of the log likelihood of $\mathcal{D}$ under $p_{\theta}$ (and furthermore, $\mathcal{L}^{\textsc{mc}}_{\textsc{nce}_k} \rightarrow \mathcal{L}_{\textsc{nce}_k}$). That is, the gradient is $0$ when the model distribution matches the empirical distribution.

\section{Negative sampling}
Negative sampling is a variation of NCE used by the popular {\tt word2vec} tool which generates a proxy corpus and also learns $\theta$ as a binary classification problem, but it defines the conditional probabilities given $(w,c)$ differently:
\begin{align*}
p(D = 0 \mid c,w) &=\frac{ 1 }{u_\theta(w,c) + 1} \\
p(D = 1 \mid c,w) &=\frac{ u_{\theta}(w,c) }{u_\theta(w,c) + 1}.
\end{align*}
This objective can be understood in several ways. First, it is equivalent to NCE when $k=|V|$ and $q$ is uniform. Second, it can be understood as the hinge objective of \newcite{collobert:2011} where the $\max$ function has been replaced with a softmax. As a result, aside from the $k=|V|$ and uniform $q$ case, the conditional probabilities of $D$ given $(w,c)$ are not consistent with the language model probabilities of $(w,c)$ and therefore the $\theta$ estimated using this as an objective will not optimize the likelihood of the language model in Eq.~\ref{eq:lm}. Thus, while negative sampling may be appropriate for word representation learning, it does not have the same asymptotic consistency guarantees that NCE has.

\section{Conclusion}
NCE is an effective way of learning parameters for an arbitrary locally normalized language model. However, negative sampling should be thought of as an alternative task for generating representations of words for use in other tasks, but not, itself, as a method for learning parameters in a generative model of language. Thus, if your goal is language modeling, you should use NCE; if your goal is word representation learning, you should consider both NCE and negative sampling.

\bibliographystyle{acl}
\bibliography{biblio}

\end{document}